\documentclass{article}
\usepackage{spconf,amsmath,amssymb,graphicx,booktabs,url}
\usepackage[utf8]{inputenc}

\title{Permutation Invariant Training of Deep Models for Speaker-Independent Multi-talker Speech Separation}
\name{Dong Yu$^1$, Morten Kolbæk$^2$, Zheng-Hua Tan$^2$, Jesper Jensen$^2$}
\address{$^1$Microsoft Research, Redmond, WA 98052, USA\\
$^2$Department of Electronic Systems, Aalborg University, Aalborg, Denmark\\
dongyu@microsoft.com, \{mok,zt,jje\}@es.aau.dk }
\begin{document}
\ninept
\maketitle
\begin{abstract}
We propose a novel deep learning training criterion, named permutation invariant training (PIT), for speaker independent multi-talker speech separation, commonly known as the cocktail-party problem. Different from the multi-class regression technique and the deep clustering (DPCL) technique, our novel approach minimizes the separation error directly. This strategy effectively solves the long-lasting label permutation problem, that has prevented progress on deep learning based techniques for speech separation. We evaluated PIT on the WSJ0 and Danish mixed-speech separation tasks and found that it compares favorably to non-negative matrix factorization (NMF), computational auditory scene analysis (CASA), and DPCL and generalizes well over unseen speakers and languages. Since PIT is simple to implement and can be easily integrated and combined with other advanced techniques, we believe improvements built upon PIT can eventually solve the cocktail-party problem.
\end{abstract}
\section{Introduction}\label{sec:intro}

Despite the significant progress made in dictating single-speaker speech in the recent years \cite{PretrainVSFineTune-Yu2010,CD-DNN-HMM-dahl2012,CD-DNN-HMM-SWB-seide2011,DNN4ASR-hinton2012}, the progress made in multi-talker mixed speech separation and recognition, often referred to as the cocktail-party problem \cite{Cherry53,AuditorySceneAnalysis-bregman1994}, has been less impressive. Although human listeners can easily perceive separate sources in an acoustic mixture, the same task seems to be extremely difficult for automatic computing systems, especially when only a single microphone recording of the mixed-speech is available \cite{MonauralSpeechSepChallenge-Cooke2010,SingleChannelSep-Weng2015}.

Nevertheless, solving the cocktail-party problem is critical to enable scenarios such as automatic meeting transcription, automatic captioning for audio/video recordings (e.g., YouTube), and multi-party human-machine interactions (e.g., in the world of Internet of things (IoT)), where speech overlapping is commonly observed.

Over the decades, many attempts have been made to attack this problem. Before the deep learning era, the most popular technique was computational auditory scene analysis (CASA) \cite{CASA-cooke2005,CASA-ellis1996}. In this approach, certain segmentation rules based on perceptual grouping cues \cite{PerceptualCuesInCASA-wertheimer1938} are (often semi-manually) designed to operate on low-level features to estimate a time-frequency mask that isolates the signal components belonging to different speakers. This mask is then used to reconstruct the signal. Non-negative matrix factorization (NMF) \cite{sparseNMF-schmidt2006,NMF-SpeechSep-smaragdis2007,SparseNMF-le2015} is another popular technique which aims to learn a set of non-negative bases that can be used to estimate mixing factors during evaluation. Both CASA and NMF led to very limited success in separating sources in multi-talker mixed speech \cite{MonauralSpeechSepChallenge-Cooke2010}. The most successful technique before the deep learning era is the model based approach \cite{IBM-SuperHuman-kristjansson2006,SpeechSepWithFactorialHMM-virtanen2006,SpeechSepWithAdaptedModel-weiss2007}, such as factorial GMM-HMM \cite{FactorialHMM-ghahramani1997}, that models the interaction between the target and competing speech signals and their temporal dynamics. Unfortunately this model assumes and only works under closed-set speaker condition. 

Motivated by the success of deep learning techniques in single-talker ASR \cite{PretrainVSFineTune-Yu2010,CD-DNN-HMM-dahl2012,CD-DNN-HMM-SWB-seide2011,DNN4ASR-hinton2012}, researchers have developed many deep learning techniques for speech separation in recent years. Typically, networks are trained based on parallel sets of mixtures and their constituent target sources \cite{SpeechSepTrainingTargets-wang2014,SpeechEnhanceWithDNN-xu2014,SpeechSepWithLSTM-weninger2015,JointMaskDNN-Huang2015}. The networks are optimized to predict the source belonging to the target class, usually for each time-frequency bin. Unfortunately, these works often focus on, and only work for, separating speech from (often challenging) background noise (or music) because speech has very different characteristics than noise/music. Note that there are indeed works that are aiming at separating multi-talker mixed speech (e.g., \cite{JointMaskDNN-Huang2015}). However, these works rely on speaker-dependent models by assuming that the (often few) target speakers are known during training.

The difficulty in speaker-independent multi-talker speech separation comes from the label ambiguity or permutation problem (which will be described in Section~\ref{sec:problem}). Only two deep learning based works \cite{SingleChannelSep-Weng2015,DeepClustering-hershey2015,DeepClustering2-isik2016} have tried to address and solve this harder problem. In Weng et al. \cite{SingleChannelSep-Weng2015}, which achieved the best result on the dataset used in 2006 monaural speech separation and recognition challenge \cite{MonauralSpeechSepChallenge-Cooke2010}, the instantaneous energy was used to solve the label ambiguity problem and a two-speaker joint-decoder with speaker switching penalty was used to separate and trace speakers. This approach tightly couples with the decoder and is difficult to scale up to more than two speakers due to the way labels are determined. Hershey et al. \cite{DeepClustering-hershey2015,DeepClustering2-isik2016} made significant progress with their deep clustering (DPCL) technique. In their work, they trained an embedding for each time-frequency bin to optimize a segmentation (clustering) criterion. During evaluation, each time-frequency bin was first mapped into the embedding space upon which a clustering algorithm was used to generate a partition of the time-frequency bins. Impressively, their systems trained on two-talker mixed-speech perform well on three-talker mixed-speech. However, in their approach it is assumed that each time-frequency bin belongs to only one speaker (i.e., a partition) due to the clustering step. Although this is often a good approximation, it is known to be sub-optimal. Furthermore, their approach is hard to combine with other techniques such as complex-domain separation.

In this paper, we propose a novel training criterion, named permutation invariant training (PIT), for speaker independent multi-talker speech separation. Most prior arts treat speech separation as either a multi-class regression problem or a segmentation (or clustering) problem. PIT, however, considers it a \emph{separation} problem (as it should be) by minimizing the separation error. More specifically, PIT first determines the best output-target assignment and then minimizes the error given the assignment. This strategy, which is directly implemented inside the network structure, elegantly solves the long-lasting label permutation problem that has prevented progress on deep learning based techniques for speech separation.

We evaluated PIT on the WSJ0 and Danish mixed-speech separation tasks. Experimental results indicate that PIT compares favorably to NMF, CASA, and DPCL and generalizes well over unseen speakers and languages. In other words, through the training process PIT learns acoustic cues for source separation, which are both speaker and language independent, similar to humans. Since PIT is simple to implement and can be easily integrated and combined with other advanced techniques we believe improvements built upon PIT can eventually solve the cocktail-party problem.

\section{Monaural speech separation}\label{sec:problem}

The goal of monaural speech separation is to estimate the individual source signals in a linearly mixed, single-microphone signal, in which the source signals overlap in the time-frequency domain. Let us denote the \(S\) source signal sequences in the time domain as \(\mathbf{x}_s(t), s=1,\cdots,S\) and the mixed signal sequence as \(\mathbf{y}(t)=\sum_{s=1}^{S} \mathbf{x}_s(t)\). The corresponding short-time Fourier transformation (STFT) of these signals are \(\mathbf{X}_s(t,f)\) and  \(\mathbf{Y}(t,f)=\sum_{s=1}^{S} \mathbf{X}_s(t,f)\), respectively, for each time \(t\) and frequency \(f\). Given \(\mathbf{Y}(t,f)\), the goal of monaural speech separation is to recover each source \(\mathbf{X}_s(t,f)\). 

In a typical setup, it is assumed that only STFT magnitude spectra is available. The phase information is ignored during the separation process and is used only when recovering the time domain waveforms of the sources.

Obviously, given only the magnitude of the mixed spectrum \(|\mathbf{Y}(t,f)|\), the problem of recovering \(|\mathbf{X}_s(t,f)|\) is ill-posed, as there are an infinite number of possible \(|\mathbf{X}_s(t,f)|\) combinations that lead to the same \(|\mathbf{Y}(t,f)|\). To overcome this core problem, the system has to learn from some training set \(\mathbb{S}\) that contains pairs of \(|\mathbf{Y}(t,f)|\) and \(|\mathbf{X}_s(t,f)|\) to look for regularities. More specifically, we train a deep learning model \(g(\cdot)\) such that \(g\left(f(|\mathbf{Y}|\right);\theta)={|\tilde{\mathbf{X}}_s|, s=1,\cdots,S}\), where \(\theta\) is a model parameter vector, and \(f(|\mathbf{Y}|)\) is some feature representation of \(|\mathbf{Y}|\). For simplicity and clarity we have omitted, and will continue to omit, time-frequency indexes when there is no ambiguity.

It is well-known (e.g., \cite{SpeechSepTrainingTargets-wang2014}) that better results can be achieved if, instead of estimating \(|\mathbf{X}_s|\) directly, we first estimate a set of masks \(\mathbf{M}_s(t,f)\) using a deep learning model \(h\left(f(|\mathbf{Y}|);\theta \right)=\tilde{\mathbf{M}}_s(t,f)\) with the constraint that \(\tilde{\mathbf{M}}_s(t,f) \geq 0\) and \(\sum_{s=1}^S \tilde{\mathbf{M}}_s(t,f) = 1\)  for all time-frequency bins \((t,f)\). This constraint can be easily satisfied with the softmax operation. We then estimate \(|\mathbf{X}_s|\) as \(|\tilde{\mathbf{X}}_s| = \tilde{\mathbf{M}}_s \circ |\mathbf{Y}| \), where \( \circ\) is the element-wise product of two operands. This strategy is adopted in this study.

Note that since we first estimate masks, the model parameters can be optimized to minimize the mean square error (MSE) between the estimated mask \(\tilde{\mathbf{M}}_s\) and the ideal ratio mask (IRM) \(\mathbf{M}_s=\frac{|\mathbf{X}_s|}{|\mathbf{Y}|}\),
\[J_m=\frac{1}{T \times F \times S}\sum_{s=1}^S \|\tilde{\mathbf{M}}_s - \mathbf{M}_s\|^2,\] 
where $T$ and $F$ denote the number of time frames and frequency bins, respectively. This approach comes with two problems. First, in silence segments, \(|\mathbf{X}_s|=0\) and \(|\mathbf{Y}|=0\), so that \(\mathbf{M}_s\) is not well defined. Second, what we really care about is the error between the estimated magnitude and the true magnitude of each source, while a smaller error on masks may not lead to a smaller error on magnitude. 

To overcome these limitations, recent works \cite{SpeechSepTrainingTargets-wang2014} directly minimize the mean squared error (MSE) 
\[J_x=\frac{1}{T \times F \times S}\sum_{s=1}^S \|\tilde{|\mathbf{X}}_s| - |\mathbf{X}_s|\|^2\] 
between the estimated magnitude and the true magnitude. Note that in silence segments \(|\mathbf{X}_s|=0\) and \(|\mathbf{Y}|=0\), and so the accuracy of mask estimation does not affect the training criterion for those segments. In this study, we estimate masks $\tilde{\mathbf{M}}_s$ which minimize \(J_x\).

\section{Permutation invariant training}\label{sec:train}
%
%

\begin{figure}[ht]
  \centering
   \includegraphics[width=1.0\linewidth]{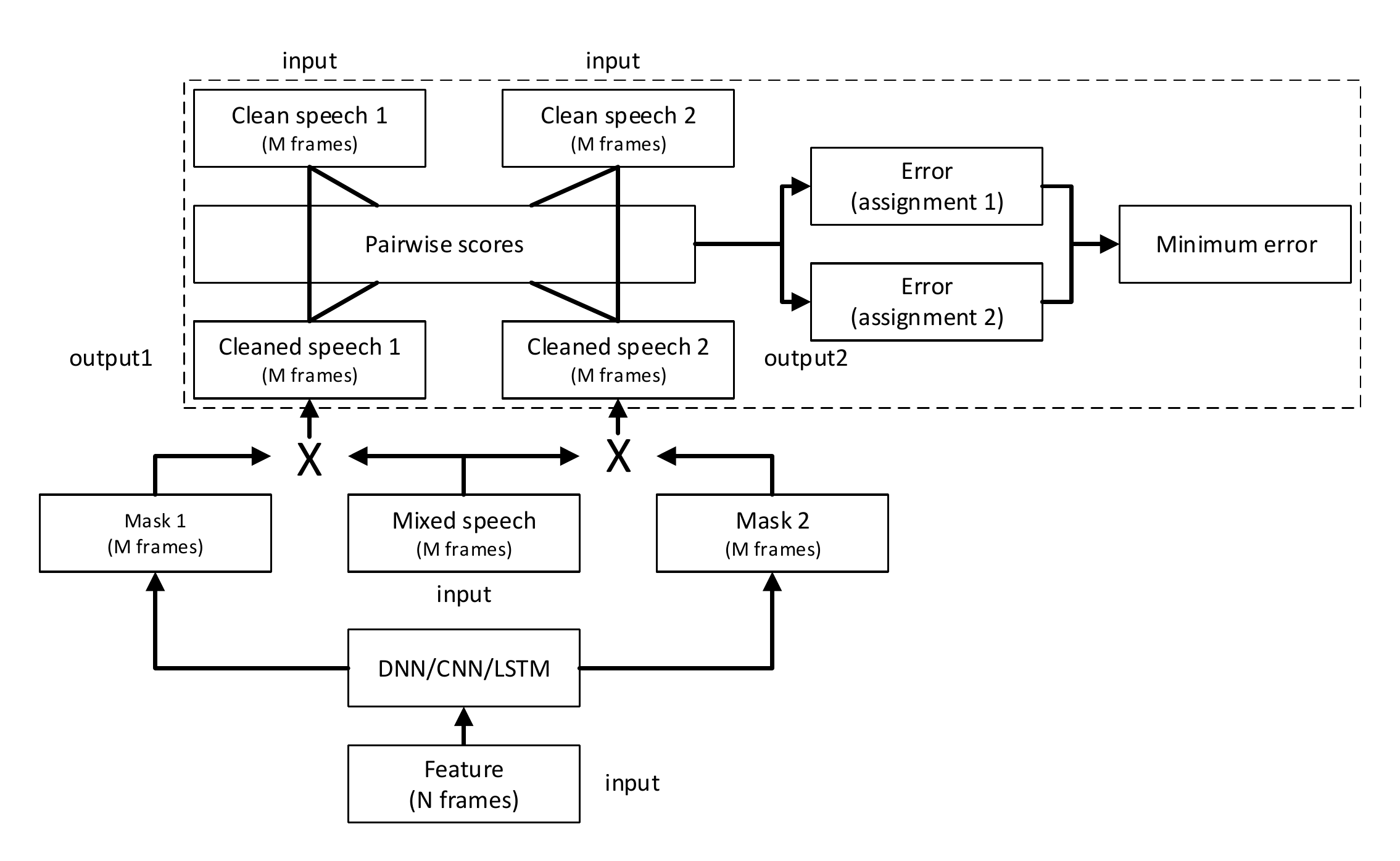}
  \caption{The two-talker speech separation model with permutation invariant training.}\label{fig:model}
\end{figure}

Except DPCL \cite{DeepClustering-hershey2015,DeepClustering2-isik2016}, all other recent speech separation works treat the separation problem as a multi-class regression problem. In their architecture, \(N\) frames of feature vectors of the mixed signal \(|\mathbf{Y}|\) are used as the input to some deep learning models, such as deep neural networks (DNNs), convolutional neural networks (CNNs), and long short-term memory (LSTM) recurrent neural networks (RNNs), to generate one (often the center) frame of masks for each talker. These masks are then used to construct one frame of single-source speech \(|\tilde {\mathbf{X}}_1|\) and \(|\tilde {\mathbf{X}}_2|\), for source 1 and 2, respectively. 

During training we need to provide the correct reference (or target) magnitude \(|\mathbf{X}_1|\) and \(|\mathbf{X}_2|\) to the corresponding output layers for supervision. Since the model has multiple output layers, one for each mixing source, and they depend on the same input mixture, reference assigning can be tricky especially if the training set contains many utterances spoken by many speakers. This problem is referred to as the label ambiguity (or permutation) problem in \cite{SingleChannelSep-Weng2015,DeepClustering-hershey2015}. Due to this problem, prior arts perform poorly on speaker-independent multi-talker speech separation. It was believed that speaker-independent multi-talker speech separation is not feasible \cite{SpeechSepTutorial-wang2016}.

The solution proposed in this work is illustrated in Figure~\ref{fig:model}. There are two key inventions in this novel model: permutation invariant training (PIT) and segment-based decision making. 

In our new model the reference source streams are given as a set instead of an ordered list. In other words, the same training result is obtained, no matter in which order these sources are listed. This behavior is achieved with PIT highlighted inside the dashed rectangular in Figure~\ref{fig:model}.  
In order to associate references to the output layers, we first determine the (total number of \(S!\)) possible assignments between the references and the estimated sources. %
We then compute the total MSE for each assignment, which is defined as the combined pairwise MSE between each reference \(|\mathbf{X}_s|\) and the estimated source \(|\tilde {\mathbf{X}}_s|\). 
The assignment with the least total MSE is chosen and the model is optimized to reduce this particular MSE. 
In other words we simultaneously conduct label assignment and error evaluation. Similar to the prior arts, PIT uses as input \(N\) successive frames (i.e., an input {\em meta-frame}) of features to exploit the contextual information. Different from the prior arts, the output of the PIT is also a window of frames. With PIT, we directly minimize the separation error at the meta-frame level. 
Although the number of speaker assignments is factorial in the number of speakers, the pairwise MSE computation is only quadratic, and more importantly the MSE computation can be completely ignored during evaluation.

During inference,  the only information available is the mixed speech. Speech separation can be directly carried out for each input meta-frame, for which an output meta-frame with \(M\) frames of speech is estimated for each stream. The input meta-frame is then shifted by one or more frames. Due to the PIT training criterion, output-to-speaker assignment may change across frames. In the simplest setup, we can just assume they do not change when reconstructing sources. Better performance may be achieved if a speaker-tracing algorithm is applied on top of the output of the network.

Once the relationship between the outputs and source streams are determined for each output meta-frame, the separated speech can be estimated, taking into account all meta-frames by, for example, averaging the same frame across meta-frames. 

\section{Experimental results}
\label{sec:exp}

\subsection{Datasets}
\label{subsec:datasets}

We evaluated PIT on the WSJ0-2mix and Danish-2mix datasets. The WSJ0-2mix dataset was introduced in \cite{DeepClustering-hershey2015} and was derived from WSJ0 corpus \cite{wsj0}. The 30h training set and the 10h validation set contains two-speaker mixtures generated by randomly selecting speakers and utterances from the WSJ0 training set si\_tr\_s, and mixing them at various signal-to-noise ratios (SNRs) uniformly chosen between 0 dB and 5 dB. The 5h test set was similarly generated using utterances from 16 speakers from the WSJ0 validation set si\_dt\_05 and evaluation set si\_et\_05.

The Danish-2mix dataset was constructed from the Danish corpus \cite{dkCorpus}, which consists of approximately 560 speakers each speaking 312 utterances with average utterance duration of approximately 5 sec. The dataset was constructed by randomly selecting a set of 45 male and 45 female speakers from the corpus, and then allocating 232, 40, and 40 utterances from each speaker to generate mixed speech in the training, validation and closed-condition\,(CC) (seen speaker) test set, respectively. 40 utterances from each of another 45 male and 45 female speakers were randomly selected to construct the open-condition\,(OC) (unseen speaker) test set. Speech mixtures were constructed in the way similar to the WSJ0-2mix dataset, but all mixed with 0 dB - the hardest condition. We constructed 10k and 1k mixtures in total in the training and validation set, respectively, and 1k mixtures for each of the CC and OC test sets. The Danish-3mix (three-talker mixed speech) dataset was constructed similarly.

In this study we focus on the WSJ0-2mix dataset so that we can directly compare PIT with published state-of-the-art results obtained using other techniques.

\subsection{Models}

Our models were implemented using the Microsoft Cognitive Toolkit (CNTK) \cite{CNTK2014}. The feed-forward DNN (denoted as DNN) has three hidden layers each with 1024 ReLU units. In (inChannel, outChannel)-(strideW, strideH) format, the CNN model has one $(1,64)-(2,2)$, four $(64,64)-(1,1)$, one $(64,128)-(2,2)$, two $(128,128)-(1,1)$, one $(128,256)-(2,2)$, and two $(256,256)-(1,1)$ convolution layers with $3 \times 3$ kernels, a pooling layer and a 1024-unit ReLU layer. The input to the models is the stack (over multiple frames) of the 257-dim STFT spectral magnitude of the speech mixture, computed using STFT with a frame size of 32ms and 16ms shift. There are $S$ output streams for $S$-talker mixed speech. Each output stream has a dimension of $257 \times M$, where $M$ is the number of frames in the output meta-frame. In our study, the validation set is only used to control the learning rate.

\subsection{Training behavior}

\begin{figure}[ht] 
  \centering
   \centerline{\includegraphics[width=1\linewidth]{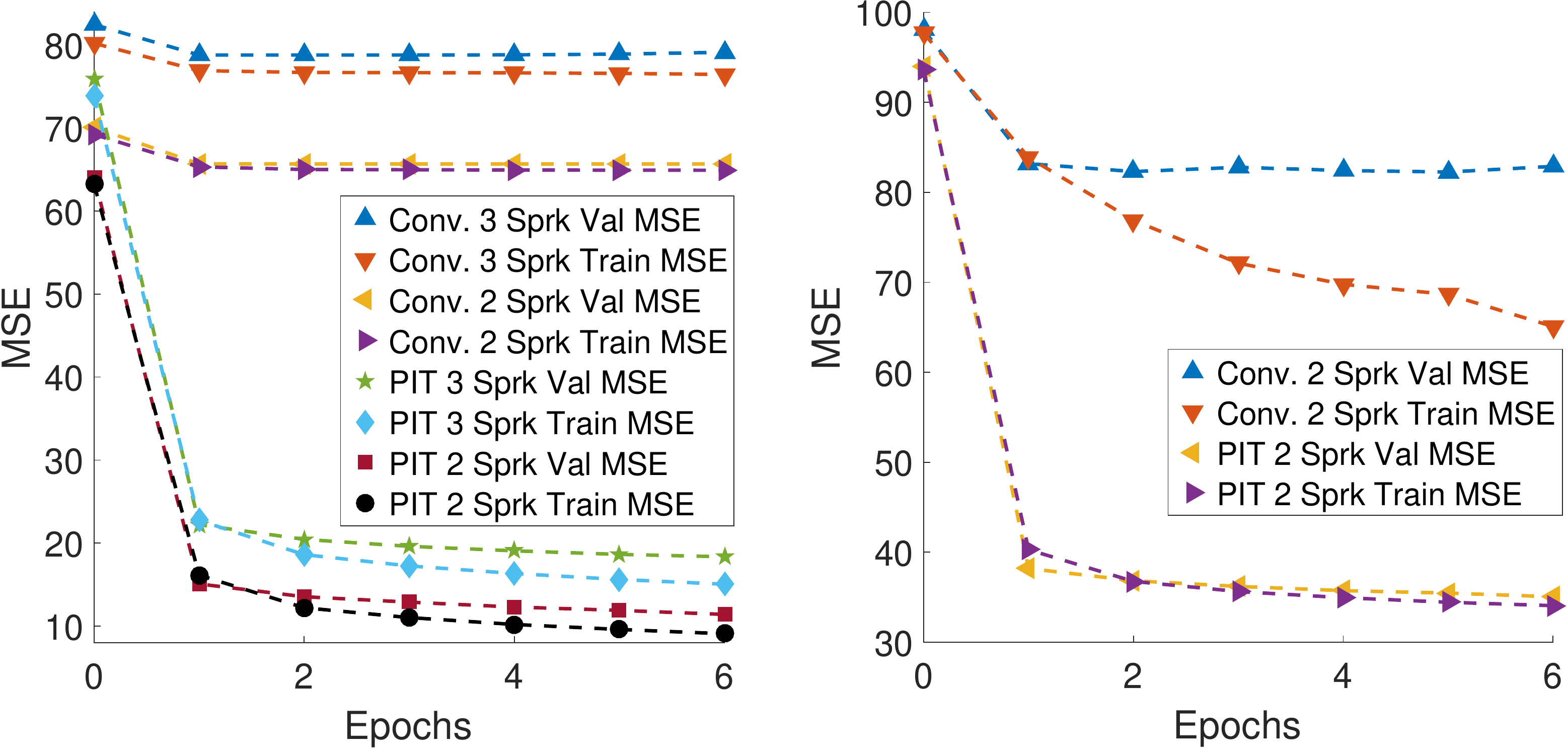}}
  \caption{MSE over epochs on the  Danish (left) and WSJ0 (right) training and validation sets with conventional training and PIT.}\label{fig:mse}
\end{figure}

In Figure~\ref{fig:mse} we plotted the DNN training progress as measured by the MSE on the training and validation set with conventional training and PIT on the mixed speech datasets described in subsection \ref{subsec:datasets}. From the figure we can see clearly that the validation MSE hardly decreases with the conventional approach due to the label permutation problem discussed in \cite{SingleChannelSep-Weng2015,DeepClustering-hershey2015}. In contrast, training converges quickly to a much better MSE for both two- and three-talker mixed speech when PIT is used.

\subsection{Signal-to-distortion ratio improvement}

We evaluated PIT on its potential to improve the signal-to-distortion ratio (SDR) \cite{vincent_performance_2006}, a metric widely used to evaluate speech enhancement performance.
\begin{table}[t]
\caption{SDR improvements (dB) for different separation methods on the WSJ0-2mix dataset.}
\label{tab:WSJ0-2mix}
\centering
\begin{tabular}{l|c|cc|cc}
\toprule
Method & Input\textbackslash Output & \multicolumn{2}{c|} {Opt. Assign} & \multicolumn{2}{c} {Def. Assign}\\ 
       &  window                    		& CC & OC		&  CC 	& OC	\\
\midrule
Oracle NMF \cite{DeepClustering-hershey2015}	& -	& -	& -		& 5.1  & -     \\ 
CASA \cite{DeepClustering-hershey2015}   	& -  & - & - 	& 2.9  	& 3.1   \\
DPCL \cite{DeepClustering-hershey2015} & 100\textbackslash100 & 6.5 & 6.5& 5.9  	& 5.8   \\
DPCL+ \cite{DeepClustering2-isik2016}   & 100\textbackslash100 & -	& -	&  -  	& 10.3     \\ 
\midrule
PIT-DNN    & 101\textbackslash101    		& 6.2 	&     6.0  	&     5.3 	&     5.2     \\ 
PIT-DNN    & 51\textbackslash51    			& 7.3 	&     7.2  	& \bf{5.7} 	& \bf{5.6}     \\ 
PIT-DNN    & 41\textbackslash7    			& 10.1 	&     10.0  &    -0.3 	&    -0.6     \\ 
PIT-DNN    & 41\textbackslash5    		& \bf{10.5} & \bf{10.4}  &   -0.6 	&    -0.8     \\ 
\midrule
PIT-CNN    & 101\textbackslash101   		& 8.4 	&     8.6  	& \bf{7.7} 	& \bf{7.8}     \\ 
PIT-CNN    & 51\textbackslash51      		& 9.6 	&     9.7  	&     7.5 	&     7.7     \\ 
PIT-CNN    & 41\textbackslash7     			& 10.7 	&     10.7  &    -0.6 	&    -0.7     \\ 
PIT-CNN    & 41\textbackslash5      	& \bf{10.9} & \bf{10.9}  &   -0.8 	&    -0.9     \\ 
\midrule
IRM  		& -     						& 12.3 	& 12.5 	& 12.3 	& 12.5     \\ 
\bottomrule
\end{tabular}
\vspace{-4mm}
\end{table}

In Table~\ref{tab:WSJ0-2mix} we summarized the SDR improvement in dB from different separation configurations for two-talker mixed speech in closed condition\,(CC) and open condition\,(OC). In these experiments each frame was reconstructed by averaging over all output meta-frames that contain the same frame. In the default assignment setup it is assumed that there is no output-speaker switch across frames (which is not true). This is the improvement achievable using PIT without any speaker tracing. In the optimal assignment setup, the output-speaker assignment for each output meta-frame is  determined based on mixing streams. This reflects the separation performance within each segment (meta-frame) and is the improvement achievable when the speakers are correctly traced. 
The gap between these two values indicates the contribution from speaker tracing. 
As a reference, we also provided the IRM result which is the oracle and upper bound achievable on this task. 

From the table we can make several observations. First, without speaker tracing (def. assign) PIT can achieve similar and better performance than the original DPCL \cite{DeepClustering-hershey2015}, respectively, with DNN and CNN, but under-performs the more complicated DPCL+ \cite{DeepClustering2-isik2016}. Note that, PIT is much simpler than even the original (simpler) DPCL and we did not fine-tune architectures and learning procedures as done in \cite{DeepClustering2-isik2016}. Second, as we reduce the output window size we can improve the separation performance within each window and achieve better SDR improvement if speakers are correctly traced (opt. assign). However, when output window size is reduced, the output-speaker assignment changes more frequently as indicated by the poor default assignment performance. Speaker tracing thus becomes more important given the larger gap between the opt. assign and def. assign. Fourth, PIT generalizes well on unseen speakers since the performances on the open and closed conditions are very close. Fifth, powerful models such as CNN consistently outperforms DNNs but the gain diminishes when the output window size is small.

\begin{table}[t]
\caption{SDR improvements (dB) based on optimal assignment for DNNs trained with Danish-2mix.}
\label{tab:twotalker}
\centering
\begin{tabular}{ccccccccc}
\toprule
Method & \begin{tabular}[c]{@{}c@{}}Input\textbackslash Output\\ window\end{tabular} & CC &  OC & \begin{tabular}[c]{@{}c@{}}WSJ0\\ OC\end{tabular} \\ 
 \midrule
IRM	   &  -						    & 17.2      	& 17.3  & 13.2  \\ 
\midrule
PIT-DNN    & 101\textbackslash101       & 9.00   	  & 8.61 	& 4.29  \\ 
PIT-DNN    & 61\textbackslash61         & 9.87   	  & 9.44  & 5.17  \\ 
PIT-DNN    & 31\textbackslash31         & 11.1   	  & 10.7  & 6.18  \\ 
PIT-DNN    & 31\textbackslash7          & 14.0   	  & 13.8  & 9.03  \\
PIT-DNN    & 31\textbackslash5          & 14.1   	  & 13.9	& 9.29  \\ 
\bottomrule
\end{tabular}
\vspace{-4mm}
\end{table}

In Table~\ref{tab:twotalker} we summarized the SDR improvement in dB with optimal assignment from different  configurations for DNNs trained on Danish-2mix. We also report SDR improvement using a dataset constructed identical to Danish-2mix but based on the si\_tr\_s data from WSJ0. Besides the findings obtained in Table~\ref{tab:WSJ0-2mix}, an interesting observation is that although the system has never seen English speech, it performs remarkably well on this WSJ0 dataset when compared to the IRM (oracle) values. These results indicate that the separation ability learned with PIT generalizes well not only across speakers but also across languages.

\section{Conclusion and discussion}\label{sec:conclusion}

In this paper, we have described a novel permutation invariant training technique for speaker-independent multi-talker speech separation. To the best of our knowledge this is the first successful work that employs the separation view (and criterion) of the task\footnote{Hershey et al. \cite{DeepClustering-hershey2015} tried PIT (called permutation free training in their paper) but failed to make it work. They retried after reading the preprint of this work and now got positive results as well.}, instead of the multi-class regression or segmentation view that are used in prior arts. This is a big step towards solving the important cocktail-party problem in a real-world setup, where the set of speakers are unknown during the training time.

Our experiments on two-talker mixed speech separation tasks demonstrate that PIT trained models generalize well to unseen speakers and languages. Although our results are mainly on two-talker separation tasks, PIT can be easily and effectively extended to the three-talker case as shown in figure \ref{fig:mse}.

In this paper we focused on PIT - the key technique that enables training for the separation of multi-talker mixed speech. PIT is much simpler yet performs better than the original DPCL \cite{DeepClustering-hershey2015} that contains separate embedding and clustering stages. 

Since PIT, as a training technique, can be easily integrated and combined with other advanced techniques, it has great potential for further improvement. We believe improvements can come from work in the following areas:

First, due to the change of output-speaker assignment across frames, there is a big performance gap between the optimal output-speaker assignment and the default assignment, especially in the same-gender case and when the output window size is small. This gap can be reduced with separate speaker tracing algorithms that exploit the overlapping frames and speaker characteristics (e.g., similarity) in output meta-frames. It is also possible to train an end-to-end system in which speaker tracing is directly built into the model, e.g., by applying PIT at utterance level. We will report these results in other papers.

Second, we only explored simple DNN/CNN structures in this work. More powerful models such as bi-directional LSTMs, CNNs with deconvolution layers, or even just larger models may further improve the performance. Hyper-parameter tuning will also help and sometimes lead to significant performance gain.

Third, in this work we reconstructed source streams from spectral magnitude only. Unlike DPCL, PIT can be easily combined with reconstruction techniques that exploit complex-valued spectrum to further boost performance.

Fourth, the acoustic cues learned by the model are largely speaker and language independent. It is thus possible to train a universal speech separation model using speech in various speakers, languages, and noise conditions. 

Finally, although we focused on monaural speech separation in this work, the same technique can be deployed in the multi-channel setup and combined with techniques such as beam-forming due to its flexibility. In fact, since beam-forming and PIT separate speech using different information, they complement with each other. For example, speaker tracing may be much easier when beam-forming is available.

\section{Acknowledgment}
We would like to thank Dr. John Hershey at MERL and Zhuo Chen at Columbia University for sharing the WSJ0-2mix data list and for valuable discussions.


\vfill\pagebreak

\bibliographystyle{IEEEbib}
\bibliography{mybib}

\end{document}